\documentclass[runningheads]{llncs}

\usepackage{eccv}

\usepackage{eccvabbrv}
\usepackage{graphicx}
\usepackage{booktabs}
\usepackage[accsupp]{axessibility}
\usepackage[T1]{fontenc}
\usepackage{amsmath}
\usepackage{amssymb}
\usepackage{algorithm}
\usepackage{algorithmic}
\usepackage{tikz}
\usetikzlibrary{positioning,shapes,arrows,calc}
\usepackage{multirow}
\usepackage{subcaption}
\usepackage{hyperref}
\usepackage{xcolor}
\usepackage{orcidlink}
\usepackage{pifont}
\graphicspath{{images/}}
\usepackage{placeins} 
\begin{document}

\title{Language-Structured Relational Q-Learning for Threat-Aware Control in Safety-Critical Driving}
\titlerunning{Emergent Threat Attention and Policy Collapse}
\author{Aditya Humnabadkar\orcidlink{0000-0002-9301-393X} \and Huaizhong Zhang\orcidlink{0009-0009-8899-4081} \and
Ardhendu Behera\orcidlink{0000-0003-0276-9000}}
\authorrunning{Aditya Humnabadkar et al.}

\institute{Department of Computer Sciences, Edge Hill University, United Kingdom\\
\email{\{humnabaa, zhangh,  beheraa\}@edgehill.ac.uk}}

\maketitle

\begin{abstract}
Natural-language-based scenario generation offers an intuitive means of describing rare and complex driving interactions, yet it is still uncertain whether training with language-structured data leads to truly adaptive control policies. We propose Language-Structured Relational Q-Learning, instantiated through an Ego-Centric Relational Q-Network (ERQ-Net), which jointly learns inter-vehicle relevance and action values from dynamic traffic graphs. Language descriptions define surrounding-vehicle behaviours during training, while prompts and semantic actor roles are hidden from the policy. ERQ-Net must therefore infer threat relevance solely from observable kinematics and interactions. Across 2,500 safety-critical scenarios, language-structured training improves test success from 49--52\% to 55--58\% and increases adversary-focused attention from 1.2$\times$ to 2.1$\times$, demonstrating emergent threat awareness. However, this representational gain does not consistently translate into adaptive control: trained policies perform similarly to the best constant action, while a portfolio of simple policies solves 76\% of scenarios. We formalise this discrepancy as a recognition--control gap and show that reward reweighting and margin shaping do not eliminate the resulting policy collapse. Evaluations of realism, criticality, semantic accuracy, and transfer of state–interface representations to CARLA further highlight both the strengths and the constraints of language-structured relational policy learning in safety-critical driving scenarios.
\end{abstract}

\section{Introduction}
\label{sec:intro}
Safety validation for autonomous vehicles requires extensive testing in rare, interactive, and safety-critical situations that are underrepresented in naturalistic driving logs and costly to reproduce physically~\cite{kalra2016driving,wang2021advsim,thomas2020perception}. Simulation platforms and scenario-generation frameworks therefore play a key role in exposing autonomous-driving systems to long-tail interactions while preserving experimental control and repeatability~\cite{dosovitskiy2017carla,xu2022safebench,fremont2019scenic}. Manually specifying each scenario's actors, conditions, and triggers is laborious, whereas natural language lets a tester simply request \emph{``an aggressive vehicle cuts in ahead of the ego''} without programming every state transition.

Recent methods translate natural-language descriptions into traffic layouts, trajectories, or executable simulator programs~\cite{tan2023language,zhong2023language,zhang2024chatscene,cai2025text2scenario}. In parallel, reinforcement learning and adversarial generation create reactive traffic agents, reveal ego-policy failures, and construct safety-critical situations~\cite{ding2020learning,suo2021trafficsim,zhang2023cat,ransiek2024goose,rowe2025ctrlsim}. These advances show that language can control scenario generation and that reinforcement learning can produce interactive behaviours. However, an executable interaction does not guarantee that a downstream ego policy can recognize and respond to the behavioural structure it implies.

This distinction raises a complementary question: \emph{what does an ego policy learn when its training distribution is structured by language-specified intent, but the language prompt and semantic actor roles are unavailable during policy execution?} Existing language-conditioned traffic-generation methods focus on controllability, fidelity, realism, or safety~\cite{tan2023language,zhong2023language,zhang2024chatscene}. Graph-based driving models capture inter-vehicle relations for prediction and decision-making~\cite{gao2020vectornet,zhou2022hivt}, but rarely test whether language-defined scenario structure induces emergent behaviour in downstream policies, leaving the link between semantic specification, threat recognition, and action selection poorly understood.

We address this problem with \emph{language-structured ego-centric relational Q-learning}. Our framework converts natural-language descriptions into schema-valid traffic configurations and executable interaction programs (cut-in, sudden-braking, overtaking, and tailgating). These specifications set surrounding actors' initial states and interaction dynamics, but the ego-policy observation excludes both the original prompt and privileged actor-role labels. Thus, language structures the policy's training experience rather than serving as a direct policy input.

To learn the ego policy, we propose the \emph{Ego-Centric Relational Q-Network} (ERQ-Net), which couples dynamic traffic-graph reasoning with value-based action selection. At each timestep, the traffic scene is a graph whose nodes represent the ego and surrounding vehicles. Using graph attention~\cite{velivckovic2017graph}, ERQ-Net selectively aggregates neighbouring-vehicle information into an ego-specific relational embedding that directly parameterizes Q-values over available manoeuvres, following deep Q-networks~\cite{mnih2015human}.The graph encoder and Q-value head are jointly optimized via a temporal-difference objective, so attention is learned by its impact on ego action selection rather than as a generic scene representation.

The methodological novelty is not a new graph-attention operator or Q-learning update, but their \emph{end-to-end ego-centric coupling} in a language-structured relational decision process. ERQ-Net jointly learns: (i) which surrounding actors matter to the ego, (ii) how to encode their spatial and kinematic relations into an ego state, and (iii) how this relational state determines action values. Because semantic actor roles are hidden, any preferential attention to the adversarial vehicle must arise from observable behaviour, such as converging velocity, lateral displacement, reduced separation, or braking patterns.

Our experiments reveal both an emergent capability and a key limitation. Training on language-structured scenarios boosts success by up to 6\% over a matched random control and raises the adversary-to-background attention ratio from $1.2\times$ to $2.1\times$, showing that ERQ-Net identifies behaviourally relevant actors from kinematics alone. Yet all
seeds plateau near the best constant action, while a portfolio of simple policies solves 76\%, leaving an 18-point gap the learned controller does not exploit. The policy thus learns \emph{where the threat is} but not reliably \emph{which action it requires}.

Our contributions are:

\begin{itemize}
\item We introduce \textbf{ERQ-Net}, an ego-centric relational Q-learning framework that jointly learns actor relevance, ego-state representations, and action values from dynamic traffic graphs via a unified temporal-difference objective.
\item We develop a controlled \textbf{language-structured training protocol} where semantic intent shapes surrounding-actor interactions, while language prompts and actor-role labels remain hidden from the ego policy. 
\item We show \textbf{emergent threat-focused attention} and formalise the \emph{recognition--control gap} between relational threat recognition and adaptive action selection. 
\item We define a reproducible \textbf{policy-collapse attractor} via constant-action and policy-portfolio analyses, and show that reward reweighting and margin shaping do not prevent this failure. 
\item We evaluate scenario realism, criticality, and semantic fidelity, and study zero-shot state-interface transfer from HighwayEnv to CARLA, distinguishing it from perception-level sim-to-real transfer.
\end{itemize}

\vspace{-0.5cm}
\section{Related Work}
\label{sec:related}
\vspace{-0.2cm}
\noindent\textbf{Language-controlled traffic generation.} 
LCTGen~\cite{tan2023language} and CTG~\cite{zhong2023language} generate language-conditioned traffic layouts and trajectories. ChatScene~\cite{zhang2024chatscene} and Text2Scenario~\cite{cai2025text2scenario} translate natural-language descriptions into executable simulator content. These methods primarily evaluate generation quality, controllability, or downstream safety improvement. Our focus is different: we isolate how the \emph{structure of the generated training distribution} changes an RL policy's internal attention and eventual failure mode when language is unavailable at inference.

\noindent\textbf{Reactive and controllable driving agents.}
TrafficSim~\cite{suo2021trafficsim} learns realistic multi-agent dynamics, while CtRL-Sim~\cite{rowe2025ctrlsim} uses return-conditioned offline RL for controllable reactive agents. Graph models such as VectorNet~\cite{gao2020vectornet} and HiVT~\cite{zhou2022hivt} provide relational inductive biases for traffic reasoning. We use established graph-attention and Q-learning operators, but formulate their coupling specifically for ego-centric relational control. This controlled choice allows us to attribute behavioural differences to the language-structured curriculum. 

\noindent\textbf{Safety-critical scenario generation.}
AdvSim~\cite{wang2021advsim}, STRIVE~\cite{rempe2022generating}, and CAT~\cite{zhang2023cat} search for adversarial trajectories or policies that expose ego failures. GOOSE~\cite{ransiek2024goose} uses goal-conditioned RL for safety-critical scenario generation, and CRITICAL~\cite{tian2024critical} integrates critical-case generation with RL training. These methods optimize scenario challenge directly; our scenarios are specified through language and constrained interaction programs, enabling semantic control but generally lower adversarial pressure.

\noindent\textbf{Simulation and failure analysis.}
SafeBench~\cite{xu2022safebench}, CARLA~\cite{dosovitskiy2017carla}, HighwayEnv~\cite{highway-env}, Scenic~\cite{fremont2019scenic}, and ScenarioRunner~\cite{scenariorunner} support scalable testing. Our contribution is not a new simulator. Instead, we use two simulators to distinguish state-interface transfer from visual robustness and make the policy-collapse mechanism a first-class experimental result.

\begin{figure*}[!t]
\centering
\includegraphics[width=\textwidth]{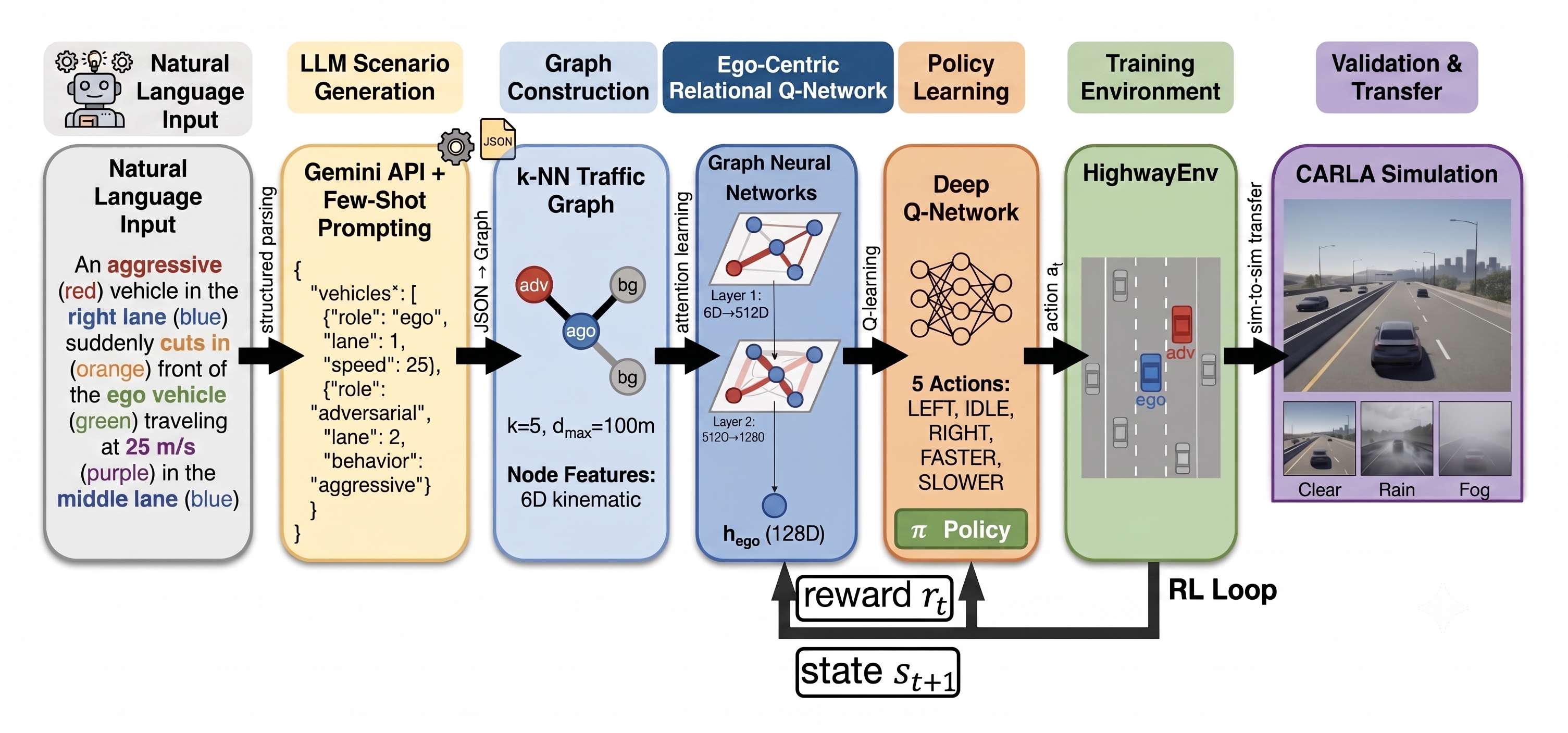}
\caption{\textbf{Language-structured relational policy-learning pipeline.} A natural-language description is parsed into schema-valid JSON and instantiated as an executable traffic interaction. At each timestep, the simulator state is represented as a dynamic k-NN traffic graph. The proposed Ego-Centric Relational Q-Network (ERQ-Net) aggregates inter-vehicle interactions into an ego-specific representation and estimates Q-values for ego manoeuvres using only observable kinematics, without language prompts, behaviour labels, or semantic actor roles. ERQ-Net is trained in HighwayEnv and evaluated in CARLA via the same graph-state interface.}
\label{fig:framework}
\vspace{-0.5cm}
\end{figure*}

\section{Problem Setting and Behavioural Diagnostics}
\label{sec:problem}
Let $\ell\in\mathcal{L}$ be a natural-language scenario description. The parser $g_{\mathrm{LLM}}$ maps $\ell$ to a structured configuration
\begin{equation}
\mathbf{c}=g_{\mathrm{LLM}}(\ell)
=\{\mathbf{c}^{\mathrm{scene}},\mathbf{c}^{\mathrm{actors}},\mathbf{c}^{\mathrm{intent}}\},
\label{eq:language_config}
\end{equation}
containing road and traffic parameters, actor initial states, and an interaction program such as cut-in, sudden braking, overtaking, or tailgating. The simulator evolves according to
\begin{equation}
s_{t+1}\sim P(\,\cdot\mid s_t,a_t;\mathbf{c}), \qquad
G_t=\Gamma(s_t), \qquad
a_t=\pi_{\boldsymbol{\phi}}(G_t),
\label{eq:closed_loop}
\end{equation}
where $\Gamma$ converts simulator state to a traffic graph and $\pi_{\boldsymbol{\phi}}$ is the ego policy. Crucially, $\mathbf{c}^{\mathrm{intent}}$ affects the actors' executed behaviour but is not included in $G_t$. The policy must infer interaction relevance from observed motion alone.

%
\subsection{Controlled Training Distributions}
\label{sec:controlled_distributions}
We train the same architecture under two distributions. $\mathcal{D}_{\mathrm{lang}}$ contains LLM-structured scenarios with a coherent adversarial interaction. $\mathcal{D}_{\mathrm{rand}}$ matches the number of scenarios and the marginal ranges of vehicle count, position, lane, and speed, but samples these factors independently and omits the structured interaction. Thus,
\begin{equation}
\pi_{\mathrm{lang}}=\operatorname{Train}(\mathcal{D}_{\mathrm{lang}}),
\qquad
\pi_{\mathrm{rand}}=\operatorname{Train}(\mathcal{D}_{\mathrm{rand}}),
\label{eq:controlled_train}
\end{equation}
with identical network, optimization, and evaluation settings. Their difference estimates the effect of language-induced scenario structure rather than a change in model capacity.

\subsection{Threat Attention and Recognition--Control Gap}
\label{sec:diagnostics}
For a rollout with adversarial actor $v_a$ and background actors $\mathcal{V}_b$, we summarise the final-layer ego attention by
\begin{equation}
A_{\mathrm{threat}}
=
\frac{\overline{\alpha}_{\mathrm{ego},a}}
{\frac{1}{|\mathcal{V}_b|}\sum_{j\in\mathcal{V}_b}
\overline{\alpha}_{\mathrm{ego},j}},
\label{eq:attention_ratio}
\end{equation}
where $\overline{\alpha}$ averages the heads and time steps after the adversarial trigger. $A_{\mathrm{threat}}>1$ indicates preferential attention to the threat.

Attention specialisation does not necessarily require adaptive control. Let $S(\pi)$ denote test success, and let
\begin{equation}
S_{\mathrm{union}}(\Pi_K)
=
\frac{1}{N}\sum_{n=1}^{N}
\mathbb{1}\!\left[\exists\,\pi_k\in\Pi_K:
\pi_k\ \text{succeeds on scenario }n\right].
\label{eq:portfolio_union}
\end{equation}
The \emph{recognition--control gap}
\begin{equation}
\Delta_{\mathrm{RC}}
=
S_{\mathrm{union}}(\Pi_K)
-
\max_{\pi_k\in\Pi_K}S(\pi_k)
\label{eq:rc_gap}
\end{equation}
measures solvable headroom that no single learned or fixed policy exploits. A large $\Delta_{\mathrm{RC}}$ alongside high $A_{\mathrm{threat}}$ indicates that the representation identifies the threat but the controller fails to condition its action on scenario type.

\subsection{Scenario Quality and Policy Objective}
\label{sec:objectives}
For rollout $s$ generated from description $\ell$, we report realism $\mathcal{R}(s)$, criticality $\mathcal{C}(s)$, and semantic fidelity $\mathcal{F}(s,\ell)$ separately, and summarize them as
\begin{equation}
\mathcal{Q}(s,\ell)=
w_R\mathcal{R}(s)+w_C\mathcal{C}(s)+w_F\mathcal{F}(s,\ell),
\qquad
(w_R,w_C,w_F)=(0.4,0.4,0.2).
\label{eq:quality}
\end{equation}
These are \emph{evaluation measures}, not the RL optimization objective. Policy learning instead uses
\begin{equation}
r_t=
\frac{v_t}{v_{\mathrm{target}}}
-\omega_{\mathrm{col}}\mathbb{1}_{\mathrm{collision}}
-\omega_{\mathrm{lc}}\mathbb{1}_{\mathrm{lane\ change}}
+r_{\mathrm{margin}}(d_{\mathrm{front}}),
\label{eq:reward}
\end{equation}
with $v_{\mathrm{target}}=30$\,m/s, $\omega_{\mathrm{col}}=1.0$, and $\omega_{\mathrm{lc}}=0.1$. The collision term is terminal and safety-oriented, while the remaining dense terms shape progress, stability, and following distance. Section~\ref{sec:collapse} shows that this apparently reasonable reward still admits a shortcut solution.
\section{Method}
\label{sec:method}

\subsection{Language-to-Scenario Instantiation}
\label{sec:llm}
A description such as \emph{``an aggressive vehicle in the right lane at 30\,m/s cuts in ahead of the ego travelling at 25\,m/s''} implicitly specifies actors, lanes, speeds, relative placement, and interaction intent. We use the Gemini API with schema-constrained few-shot prompting (system instruction, three examples, JSON schema, and user description) to extract lane index $\lambda\in\{0,1,2,3\}$, longitudinal position $p_x\in[0,1000]$\,m, speed $v\in[20,35]$\,m/s, behaviour style $\beta \in \{\text{cautious},\allowbreak \text{normal},\allowbreak \text{aggressive}\}$, role $\rho\in\{\text{ego},\text{adversarial},\text{background}\}$, and interaction type.

The parser is replaceable by any model that emits schema-valid JSON. Behaviour style and actor role configure the simulator rather than the policy input: the adversarial actor executes a cut-in, sudden-braking, overtaking, or tailgating program within a trigger distance of the ego. The GAT-DQN controls only the ego vehicle. This distinction is important: the RL policy is evaluated \emph{under} language-conditioned scenarios; it does not generate the adversarial interaction itself.

\subsection{Dynamic Traffic Graph}
\label{sec:graph}
At timestep $t$, the simulator state is converted into $G_t=(V_t,E_t)$ over $n$ vehicles. We use local k-NN connectivity
\begin{equation}
E_t=\left\{(i,j):j\in\mathcal{N}_k(i)
\land\|\mathbf{p}_i-\mathbf{p}_j\|_2<d_{\max}\right\},
\label{eq:edges}
\end{equation}
with $k=5$, $d_{\max}=100$\,m, bidirectional edges, and self-loops. Each vehicle has the observation-only feature
\begin{equation}
\mathbf{x}_i=
[\tilde p_{x,i},\tilde p_{y,i},
\tilde u_{x,i},\tilde u_{y,i},
\cos\psi_i,\sin\psi_i],
\label{eq:node_features}
\end{equation}
where positions are normalised by road dimensions, velocities by $35$\,m/s, and heading is represented continuously. Semantic role and interaction labels are deliberately excluded. Relative distance, lane offset, and time-to-collision are computed from simulator state for triggering and evaluation, but are not treated as privileged semantic inputs.

\subsection{Ego-Centric Relational Q-Network (ERQ-Net)}
\label{sec:ical}
\begin{figure}[!t]
\centering
\includegraphics[width=0.9\columnwidth]{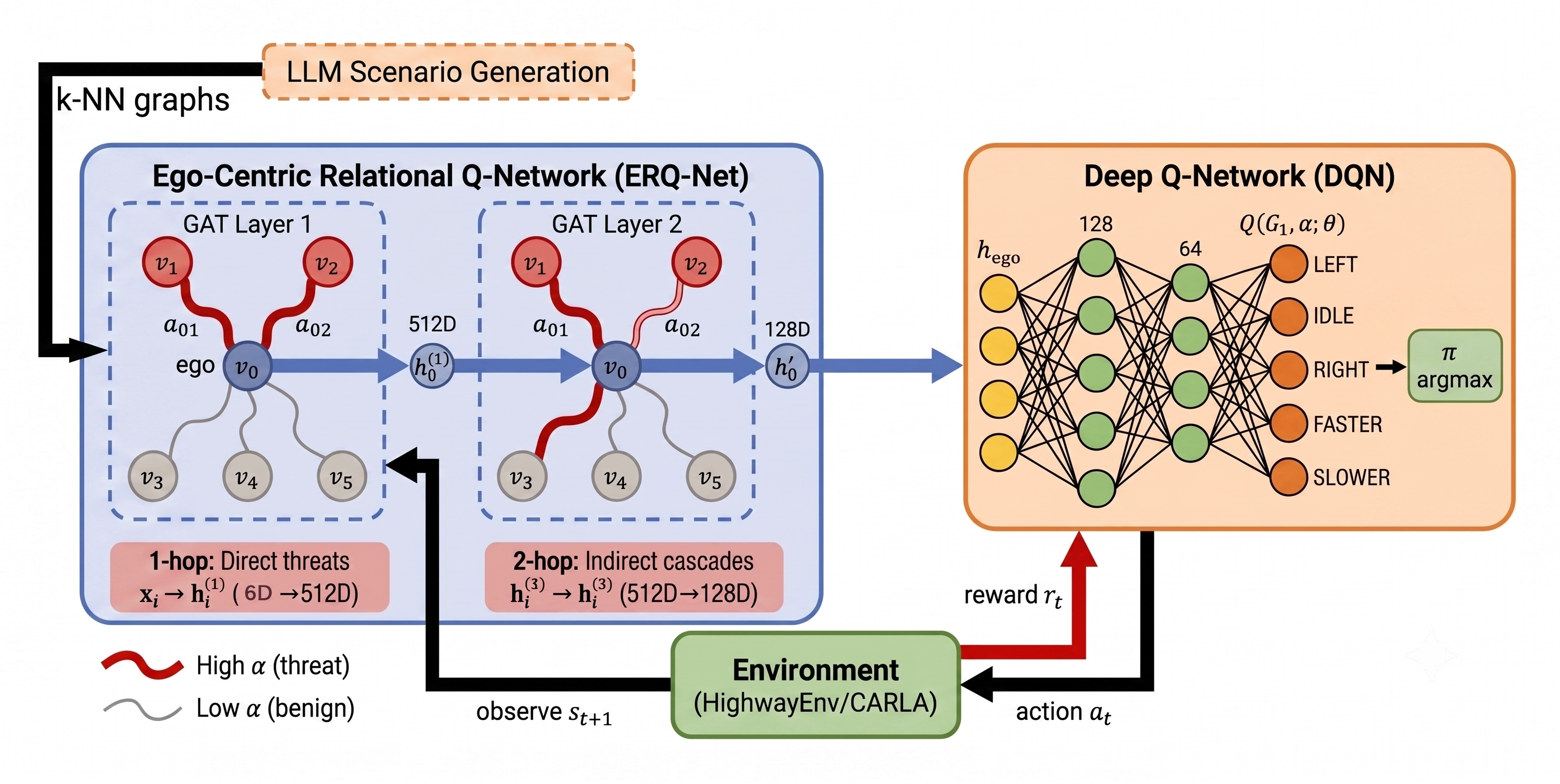}
\caption{\textbf{ERQ-Net architecture.} Multi-head graph attention aggregates local traffic interactions into an ego-centric representation that directly parameterizes Q-values for five manoeuvers, with final-layer attention serving as a diagnostic of emergent threat awareness.}
\label{fig:architecture}
\vspace{-0.5cm}
\end{figure}
We instantiate the ego policy as an ERQ-Net, which couples dynamic traffic-graph reasoning with value-based action selection. Given the traffic graph $G_t=(V_t,E_t)$, ERQ-Net learns the relevance of surrounding vehicles and constructs an ego-centric representation for control. Language prompts and semantic actor roles are excluded from the policy input; interaction relevance must therefore be inferred from observable spatial and kinematic patterns. Let $\mathbf{h}_i^{(0)}=\mathbf{x}_i$ denote the feature of vehicle $i$. For attention head $h$ at layer $c$, the normalised attention coefficient between vehicle $i$ and neighbour $j$ is 
\begin{equation} 
\alpha_{ij}^{(c,h)} = \operatorname{softmax}_{j\in\mathcal{N}_k(i)} \left[ \operatorname{LeakyReLU} \left( \left(\mathbf{a}^{(c,h)}\right)^{\top} \left[ \mathbf{W}^{(c,h)}\mathbf{h}_i^{(c)} \Vert \mathbf{W}^{(c,h)}\mathbf{h}_j^{(c)} \right] \right) \right]. 
\label{eq:erq_attention} 
\end{equation} 
The node representation is updated by aggregating information across its neighbourhood and attention heads: 
\begin{equation} 
\mathbf{h}_i^{(c+1)} = \sigma \left( \frac{1}{H} \sum_{h=1}^{H} \sum_{j\in\mathcal{N}_k(i)} \alpha_{ij}^{(c,h)} \mathbf{W}^{(c,h)} \mathbf{h}_j^{(c)} \right). 
\label{eq:erq_aggregation} 
\end{equation} 
ERQ-Net uses two graph-attention layers with four heads. 
The ego-centric state is obtained by selecting the final ego-node representation \(\mathbf{h}_t^{\mathrm{ego}}\). 
This ego-centric construction allows ERQ-Net to associate cues such as closing velocity, lateral motion, and reduced separation with interaction risk, without receiving explicit actor-role supervision. The final-layer attention coefficients are also retained for the threat-attention analysis in Eq.~\ref{eq:attention_ratio}.

\subsection{Joint Relational Q-Learning}
\label{sec:dqn}

The ego-centric representation $\mathbf{h}_t^{\mathrm{ego}}$ is mapped by a two-layer MLP
$f_{\theta_q}$ ($128\!\rightarrow\!64\!\rightarrow\!5$, ReLU) to Q-values over the set of discrete manoeuvres $
\mathcal{A}
=
\{
\texttt{LEFT}, 
\texttt{IDLE},$ 
$\texttt{RIGHT},
\texttt{FASTER},
\texttt{SLOWER}
\}$.
The resulting relational action-value function is
\begin{equation}
Q_{\boldsymbol{\phi}}(G_t,a)
=
\left[
f_{\theta_q}
\left(
\mathbf{h}_t^{\mathrm{ego}}
\right)
\right]_{a},
\qquad
\boldsymbol{\phi}
=
\{\theta_g,\theta_q\},
\label{eq:relational_q}
\end{equation}
where $\theta_g$ and $\theta_q$ denote the graph-encoder and Q-head
parameters, respectively. 
For a replay-buffer transition
$(G_t,a_t,r_t,G_{t+1},\delta_t)$, the temporal-difference target is
\begin{equation}
y_t
=
r_t
+
\gamma(1-\delta_t)
\max_{a'\in\mathcal{A}}
Q_{\boldsymbol{\phi}^{-}}(G_{t+1},a'),
\label{eq:td_target}
\end{equation}
where $\delta_t$ indicates a terminal transition and
$\boldsymbol{\phi}^{-}$ denotes the target-network parameters.
ERQ-Net is optimized using
\begin{equation}
\mathcal{L}_{\mathrm{ERQ}}(\boldsymbol{\phi})
=
\mathbb{E}_{\mathcal{B}}
\left[
\left(
y_t
-
Q_{\boldsymbol{\phi}}(G_t,a_t)
\right)^2
\right].
\label{eq:loss}
\end{equation}

Because $\theta_g$ and $\theta_q$ are jointly optimised via Eq.~\ref{eq:loss}, relational attention is learned according to its utility for ego action selection rather than through a separate representation objective. This end-to-end coupling enables ERQ-Net to jointly learn actor relevance, ego-centric state construction, and action valuation.


\subsection{State-Interface Transfer to CARLA}
\label{sec:validation}
To assess cross-simulator portability, the ERQ-Net policy trained in HighwayEnv is transferred to CARLA without fine-tuning. Simulator-specific states are mapped to the common graph representation defined in Section~\ref{sec:graph}, preserving the node features, neighbourhood construction, and action interface used during training. The same learned parameters are then applied directly in CARLA under different vehicle dynamics and road geometries. 

Because ERQ-Net operates on simulator-derived kinematics rather than visual observations, this experiment evaluates \emph{state-interface transfer}, not perception-level domain adaptation or real-world sim-to-real generalisation. Likewise, evaluations under different weather and lighting conditions measure the invariance of the kinematic graph interface rather than visual robustness.

\section{Experiments}
\label{sec:experiments}
%
\subsubsection{Scenarios and Splits:} The pipeline generates 2,500 highway scenarios specifying lane assignments, vehicle positions and velocities, behaviour styles, and one scripted safety-critical interaction. We used 2,000 scenarios for training and 500 for testing. The test set is balanced by traffic density: sparse (3--4 vehicles, 166 scenarios), moderate (5--6 vehicles, 166 scenarios) and dense (7--10 vehicles, 168 scenarios). The random-control training set matches the language-structured set in size and marginal parameter ranges, but independently samples vehicle placements and behaviours without a coherent adversarial interaction. ERQ-Net is trained in HighwayEnv~\cite{highway-env}, while cross-simulator state-interface transfer is evaluated in CARLA~\cite{dosovitskiy2017carla}.

\subsubsection{Evaluation Metrics:} Policy performance is evaluated in the same 500-scenario test set using greedy actions on three random seeds. We report collision-free completion (success), collision rate, mean episode length, and cumulative reward. The awareness of emerging threats is measured by the threat-attention ratio $A_{\mathrm{threat}}$ in Eq.~\ref{eq:attention_ratio}, while the recognition--control gap $\Delta_{\mathrm{RC}}$ follows Eq.~\ref{eq:rc_gap}.

The quality of the scenario is assessed through realism $\mathcal{R}$, criticality $\mathcal{C}$, and semantic fidelity $\mathcal{F}$. Realism is the proportion of rollout timesteps without speed-limit, road-boundary, or unsafe-proximity violations. Criticality is measured as the fraction of scenarios producing an ego collision \emph{under a fixed non-reactive reference ego} (constant velocity, no avoidance), so that $\mathcal{C}$ characterises the adversarial pressure inherent to the scenario distribution independently of the trained controller: 
\begin{equation} 
\mathcal{C} = \frac{1}{N} \sum_{n=1}^{N} \mathbb{1} \left[ \mathrm{collision}(s_n) \right], 
\label{eq:criticality} 
\end{equation} 
where $N$ is the number of scenarios and $s_n$ is the rollout for scenario $N$ under the constant-velocity reference ego. The collision definition and episode horizon are identical to those used for the trained-policy success metric, so the two are directly comparable.Semantic fidelity measures the agreement between the generated JSON specification and the realized initial configuration and interaction. We report the three components separately, together with the composite score $\mathcal{Q}$ in Eq.~\ref{eq:quality}, to avoid obscuring their individual trade-offs.

\subsubsection{Implementation Details.} ERQ-Net is implemented in PyTorch and PyTorch Geometric and trained on one NVIDIA A100 GPU. The graph encoder uses two four-head attention layers, producing a 128-D ego representation, followed by a two-layer Q-value head ($128\!\rightarrow\!64\!\rightarrow\!5$, ReLU). We train for 1,000 episodes using Adam with learning rate $3\times10^{-5}$ and discount factor $\gamma=0.99$. The replay buffer stores 10,000 transitions with a mini-batch size of 64, and the target network is synchronized every 20 episodes. Exploration follows an $\epsilon$-greedy policy, with $\epsilon$ decaying from $1.0$ to $0.05$ by a factor of $0.995$ per episode. Greedy evaluation is performed every 50 episodes across three random seeds.
\vspace{-0.3cm}
\subsection{Results}
\subsubsection{Scenario Instantiation Reliability.} Table~\ref{tab:llm_results} evaluates the reliability of the language-to-scenario stage. Here, ``success'' denotes the generation of schema-valid and executable JSON, rather than semantic correctness of the resulting rollout.
%
\begin{table}[!t]
\centering
\begin{minipage}[t]{0.44\textwidth}
\centering
\caption{Reliability of language-to-scenario instantiation.}
\label{tab:llm_results}
\begin{tabular}{lc}
\toprule
\textbf{Metric} & \textbf{Value} \\
\midrule
Total scenarios & 2,500 \\
Valid on first attempt & 94.2\% \\
Valid after one retry & 98.5\% \\
Template fallback & 1.5\% \\
Mean API latency & 1.2s \\
\bottomrule
\end{tabular}
\end{minipage}
\hfill
\begin{minipage}[t]{0.52\textwidth}
\centering
\caption{Recognition--control analysis. Although trained ERQ-Net policies outperform random and untrained agents, they remain comparable to the best constant action. The policy union identifies substantial, scenario-dependent headroom that remains unexploited.}
\label{tab:collapse}
\resizebox{\textwidth}{!}{
\begin{tabular}{lc} 
\toprule 
\textbf{Policy} & \textbf{Success} \\ 
\midrule 
Random action policy & 31--34\% \\ 
Untrained ERQ-Net (3 initializations) & 45--49\% \\ 
Best constant action (\texttt{SLOWER}) & 57\% \\ 
Trained ERQ-Net (3 seeds) & 55--58\% \\ 
\midrule 
ERQ-Net + reduced speed reward & 48--53\% \\ 
ERQ-Net + margin shaping & 48--52\% \\ 
\midrule 
Portfolio union (12 diverse policies) & \textbf{76\%} \\ 
\bottomrule 
\end{tabular}
}
\end{minipage}
\end{table}
The parser produces valid executable configurations for 94.2\% of prompts on the first attempt and 98.5\% after one retry. The remaining 1.5\% use a template fallback, ensuring that all descriptions produce executable scenarios. Semantic agreement between the specification and the actual interaction is assessed separately using the fidelity metric $\mathcal{F}$.
\subsubsection{Effect of Language-Structured Training.} 
\label{sec:ablation_intent} 
Table~\ref{tab:ablation_intent} addresses whether the semantic structure introduced by language-conditioned scenarios improves relational policy learning. The language-structured and random-control agents use the same ERQ-Net architecture, optimization settings, training-set size, and marginal parameter ranges; only the coherence of the generated interaction differs.

Language-structured training raises the best observed success from 52\% to 58\% (six points over the matched random control) and the threat-attention ratio from $1.2\times$ to $2.1\times$. Since prompts and role labels are hidden from ERQ-Net, this preferential attention is induced by coherent interaction patterns in the training distribution
rather than by inference-time supervision, supporting the emergence of threat-aware representations from behaviour alone.
\subsubsection{Relational-Encoder Ablation.} 
\label{sec:arch_baselines} 
Table~\ref{tab:baseline} evaluates whether the performance gain arises from the relational graph structure and the relevance of the learned actor. Flattening vehicle features into an MLP yields 44--47\% success. Uniform graph aggregation improves this to 48--51\%, showing a benefit from explicitly representing inter-vehicle relations.Learned attention improves performance further: the single-head GAT reaches 51--54\%, and ERQ-Net 55--58\%.

The transformer set encoder is the strongest alternative at 52--55\%, indicating that attention to surrounding vehicles is itself valuable. ERQ-Net retains a modest three-point advantage, suggesting that graph-local neighbourhood structure and multi-head interaction modelling provide an additional inductive bias. Because the ranges partially overlap across seeds, we interpret this as a consistent but moderate improvement rather than a decisive architectural margin.
\subsubsection{Training Dynamics and Policy Collapse}
\label{sec:collapse}


\begin{table}[!t]
\centering
\begin{minipage}[t]{0.50\textwidth}
\centering
\caption{Effect of language-structured training. All agents are evaluated on the same 500 language-structured test scenarios.}
\label{tab:ablation_intent}
\vspace{-.5em}
\resizebox{\textwidth}{!}{%
\begin{tabular}{lcc} 
\toprule 
\textbf{Training setting} & \textbf{Success} & \textbf{$A_{\mathrm{threat}}$} \\ 
\midrule 
Language-structured & 55--58\% & $2.1{\times}\!\pm\!0.3$ \\ 
Random-control & 49--52\% & $1.2{\times}\!\pm\!0.2$ \\ 
Untrained & 45--49\% & $1.0{\times}\!\pm\!0.1$ \\ 
\bottomrule 
\end{tabular}}
\end{minipage}
\hfill
\begin{minipage}[t]{0.4\textwidth}
\centering
\caption{Relational-encoder ablation under identical training and evaluation settings.}
\label{tab:baseline}
\vspace{-.5em}
\resizebox{\textwidth}{!}{%
\begin{tabular}{lcc} 
\toprule 
\textbf{Encoder} & \textbf{Success} & \textbf{$\Delta$} \\ 
\midrule 
MLP & 44--47\% & $-11$ \\ 
GCN & 48--51\% & $-7$ \\ 
Single-head GAT & 51--54\% & $-4$ \\ 
Transformer set encoder & 52--55\% & $-3$ \\ 
\textbf{ERQ-Net} & \textbf{55--58\%} & --- \\ 
\bottomrule 
\end{tabular}}
\end{minipage}
\vspace{-.8em}
\end{table}
Table~\ref{tab:training} shows that all seeds enter a narrow 54--58\% performance band by episode 50 and remain within it throughout training. The improvement over random actions and untrained ERQ-Net confirms that learning occurs. As Table~\ref{tab:collapse} shows, trained policies (55--58\%) remain comparable to always selecting \texttt{SLOWER} (57\%), while clearly exceeding random action (31--34\%) and untrained networks (45--49\%). This suggests a \emph{distribution-level attractor}: the Q-function learns the action with the strongest average return rather than a reliable mapping from interaction state to manoeuvre. 

This plateau is not a ceiling imposed by uniformly unsolvable scenarios.The union of 12 diverse policies which includes the trained ERQ-Net and the constant-action baselines achieves 76\%, while the best single policy in this set reaches 58\%, giving a recognition--control gap of $\Delta_{\mathrm{RC}}=76-58=18$ percentage points. Thus, language-structured training enables ERQ-Net to identify the relevant threat, but the learned Q-function does not consistently convert that representation into scenario-dependent control. Reducing the speed-reward contribution and introducing dense safety-margin shaping do not eliminate the attractor, instead reducing success to 48--53\% and 48--52\%, respectively.
\begin{table}[!t]
\centering
\begin{minipage}[t]{0.45\textwidth}
\centering
\footnotesize
\setlength{\tabcolsep}{2pt}
\caption{ERQ-Net success across three seeds on the balanced
500-scenario test set, with greedy evaluation every 50 episodes.}
\vspace{-.5em}
\label{tab:training}
\begin{tabular}{lccccc}
\toprule
\textbf{Episode} & 50 & 500 & 750 & 1000 \\
\midrule
Seed 0 & 55.0 & 55.0 & 56.0 & 54.0 \\
Seed 1 & 55.0 & 55.0 & 54.0 & 56.0 \\
Seed 2 & 55.0 & 54.0 & 58.0 & 56.0 \\
\midrule
Mean$\pm$std & 55.0{\tiny$\pm$0.0} & 54.7{\tiny$\pm$0.5}
 & 56.3{\tiny$\pm$0.9}
 & 55.3{\tiny$\pm$0.9} \\
\bottomrule
\end{tabular}
\end{minipage}
\hfill
\begin{minipage}[t]{0.46\textwidth}
\centering
\footnotesize
\setlength{\tabcolsep}{4pt}
\caption{
Decision-frequency ablation (3 seeds). The decision window is the number of ego actions between the adversarial trigger and potential impact. Success is given in percent; Gap is the shortfall in percentage points from the 76\% portfolio union in Table~\ref{tab:collapse}.
}
\vspace{-.5em}
\label{tab:freq_ablation}
\begin{tabular}{lccc}
\toprule
\textbf{Freq.} & \textbf{Decision} & \textbf{Success} & \textbf{Gap(pp)} \\
\midrule
1\,Hz & 1--2 & 55--58 & 18 \\
2\,Hz & 3--4 & 59--62 & 14 \\
5\,Hz & 8--10 & 64--68 & 8 \\
10\,Hz & 16--20 & 66--69 & 7 \\
\bottomrule
\end{tabular}
\end{minipage}
\vspace{-.8em}
\end{table}

\vspace{-0.5em}
\subsubsection{Decision-Frequency Ablation.}
\label{sec:freq_ablation}
The policy-collapse analysis in Section~\ref{sec:collapse} indicates that the 1\,Hz controller may lack the temporal resolution to turn threat recognition into an appropriate response. In the most critical interactions, the ego vehicle has only one or two action opportunities between the adversarial trigger and potential impact. This narrow decision budget may prevent the Q-function from observing the interaction’s evolution and choosing a scenario-dependent manoeuvre. Table~\ref{tab:freq_ablation} tests this hypothesis by retraining ERQ-Net at 1, 2, 5, and 10\,Hz. We fix the scenario distribution, simulator dynamics, graph architecture, reward, and effective per-second discount factor, varying only the number of policy decisions in the critical interaction window. Higher decision frequency consistently helps: success increases from 55--58\% at 1\,Hz to 59--62\% at 2\,Hz and 64--68\% at 5\,Hz. The recognition--control gap shrinks from 18 to 8 percentage points, indicating that more action opportunities let ERQ-Net better translate threat awareness into control. Raising the frequency to 10\,Hz gives only a small additional gain, to 66--69\% success and a further 1-point gap reduction.

These diminishing returns indicate that temporal resolution is important but not the only cause of policy collapse. A larger decision budget greatly improves scenario-dependent behaviour, yet the remaining 7-point gap indicates that higher-frequency control alone cannot capture the full diversity of successful responses in the policy portfolio. This finding motivates the objective and policy-level extensions in Section~\ref{sec:limitations}.

%
\vspace{-0.8em}
\subsubsection{Scenario Quality.} 
\label{sec:eq2_results}
Table~\ref{tab:eq2} reports the quality of the scenarios executed. The realism score $\mathcal{R}=0.94$ indicates that 94\% of the evaluated timesteps satisfy the specified kinematic and road-validity constraints. The criticality score is $\mathcal{C}=0.75$, indicating that 75\% of scenarios induce a collision against the non-reactive reference ego. This measure of scenario difficulty is distinct from the 55--58\% success of the trained ERQ-Net, which actively avoids collisions.The $\mathcal{F}=0.86$ indicates strong agreement between the generated JSON configuration and the placement and interaction realised. 

The resulting composite score is $\mathcal{Q}=0.85$. Shifting $0.1$ weight between realism and criticality changes it only from 0.83 to 0.87. We still report the three components separately, since the composite score reflects a chosen operational trade-off rather than a universal quality measure.
\vspace{-.8em}
\subsubsection{Contextual Comparison with CAT.}
\label{sec:cat_comparison}

\begin{table}[!t]
\centering
\begin{minipage}[t]{0.54\textwidth}
\centering
\caption{Contextual comparison with CAT~\cite{zhang2023cat}. Different simulators, initialization sources, and evaluation policies prevent a direct performance ranking.}
\label{tab:cat}
\vspace{-.8em}
\resizebox{\textwidth}{!}{%
\begin{tabular}{lcc}
\toprule
\textbf{Property} & \textbf{CAT} & \textbf{Ours} \\ 
\midrule 
Scenario source & Waymo log & Language \\ 
Language controllability & \ding{55} & \checkmark \\ 
Requires real-log initialisation & \checkmark & \ding{55} \\ 
Closed-loop execution & \checkmark & \checkmark \\ 
Generation cost per scenario & 0.198\,s & 1.2\,s \\ 
Reported realism $\mathcal{R}$ & n.r. & 0.94 \\
\bottomrule
\end{tabular}}
\end{minipage}
\hfill
\begin{minipage}[t]{0.42\textwidth}
\centering
\caption{Scenario-quality components and sensitivity of the composite score to nearby weighting choices.}
\label{tab:eq2}
\vspace{-.8em}
\resizebox{\textwidth}{!}{%
\begin{tabular}{lcccc}
\toprule
 & $\mathcal{R}$ & $\mathcal{C}$ & $\mathcal{F}$ & $\mathcal{Q}$ \\
\midrule
$(0.4,0.4,0.2)$ & 0.94 & 0.75 & 0.86 & 0.85 \\
$(0.5,0.3,0.2)$ & \multicolumn{3}{c}{(same)} & 0.87 \\
$(0.3,0.5,0.2)$ & \multicolumn{3}{c}{(same)} & 0.83 \\
\bottomrule
\end{tabular}}
\end{minipage}
\vspace{-.8em}
\end{table}
\begin{figure}[!t]
\centering

\begin{subfigure}[b]{.9\columnwidth}
    \includegraphics[width=\textwidth]{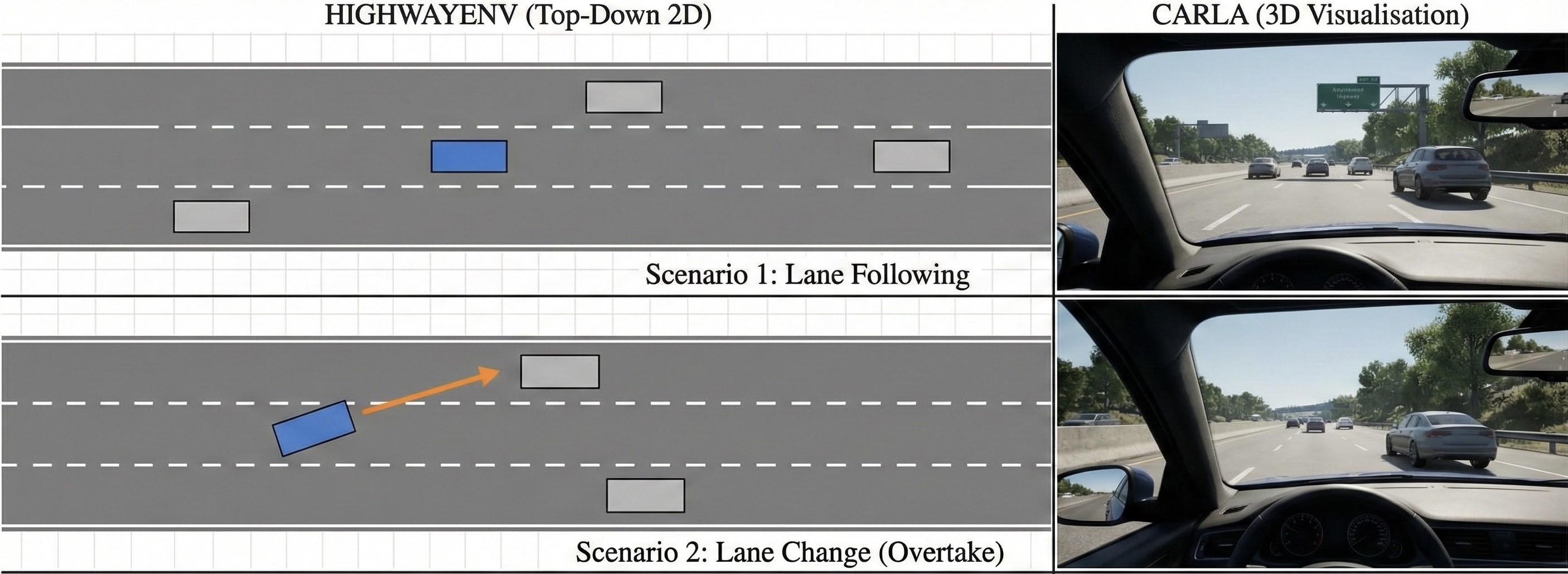}
    \caption{Lane following and overtaking scenarios. \textbf{Prompt:} \textit{"Ego vehicle following/lane changing on a straight highway" }}
    \label{fig:transfer_main}
\end{subfigure}

\vspace{0.2cm}

\begin{subfigure}[b]{0.5\columnwidth}
    \includegraphics[width=\textwidth]{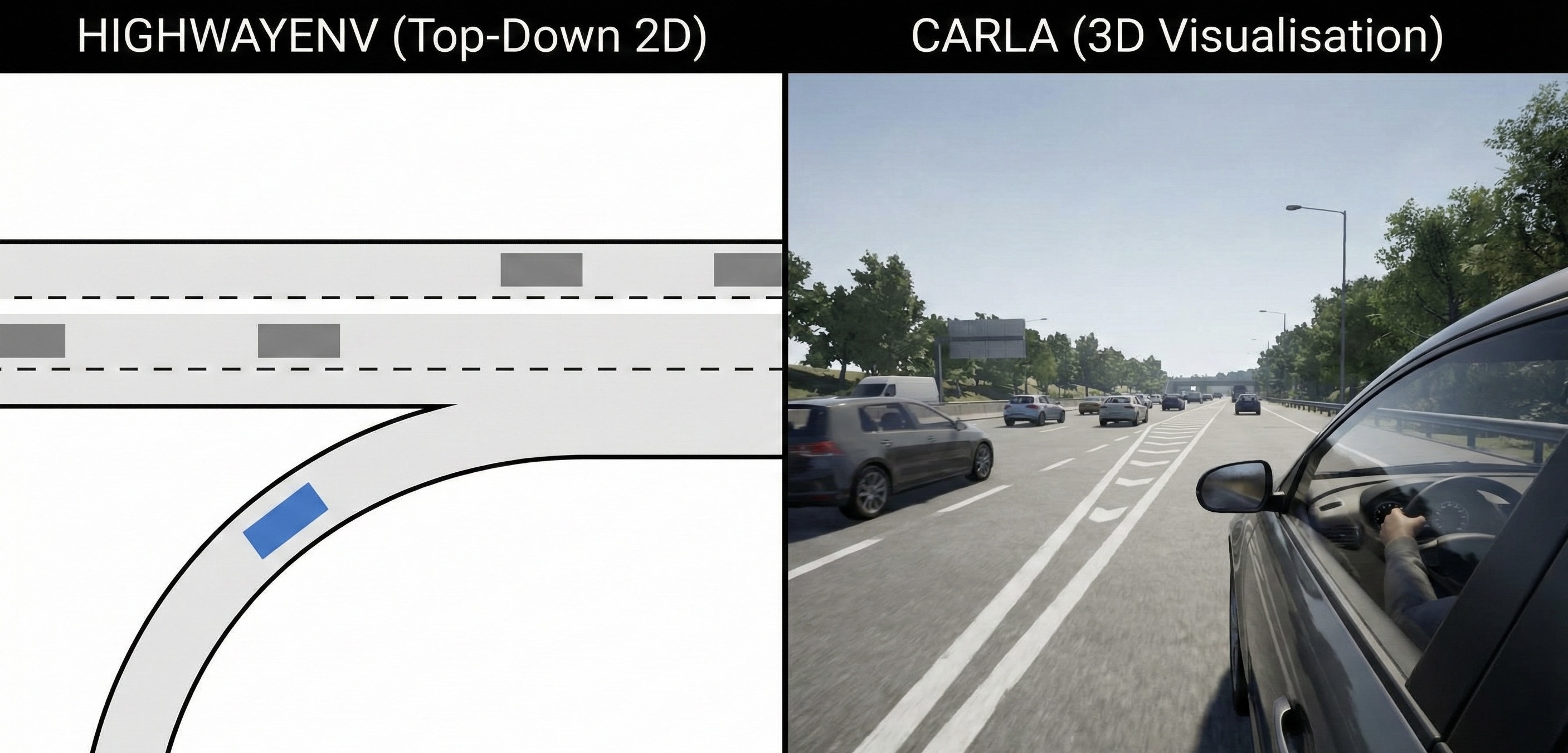}
    \caption{On-ramp merging.
    \textbf{Prompt:} \textit{"ego vehicle performs on-ramp merging with curved geometry into highway traffic"}}
    \label{fig:scenario_ramp}
\end{subfigure}
\hfill
\begin{subfigure}[b]{0.49\columnwidth}
    \includegraphics[width=\textwidth]{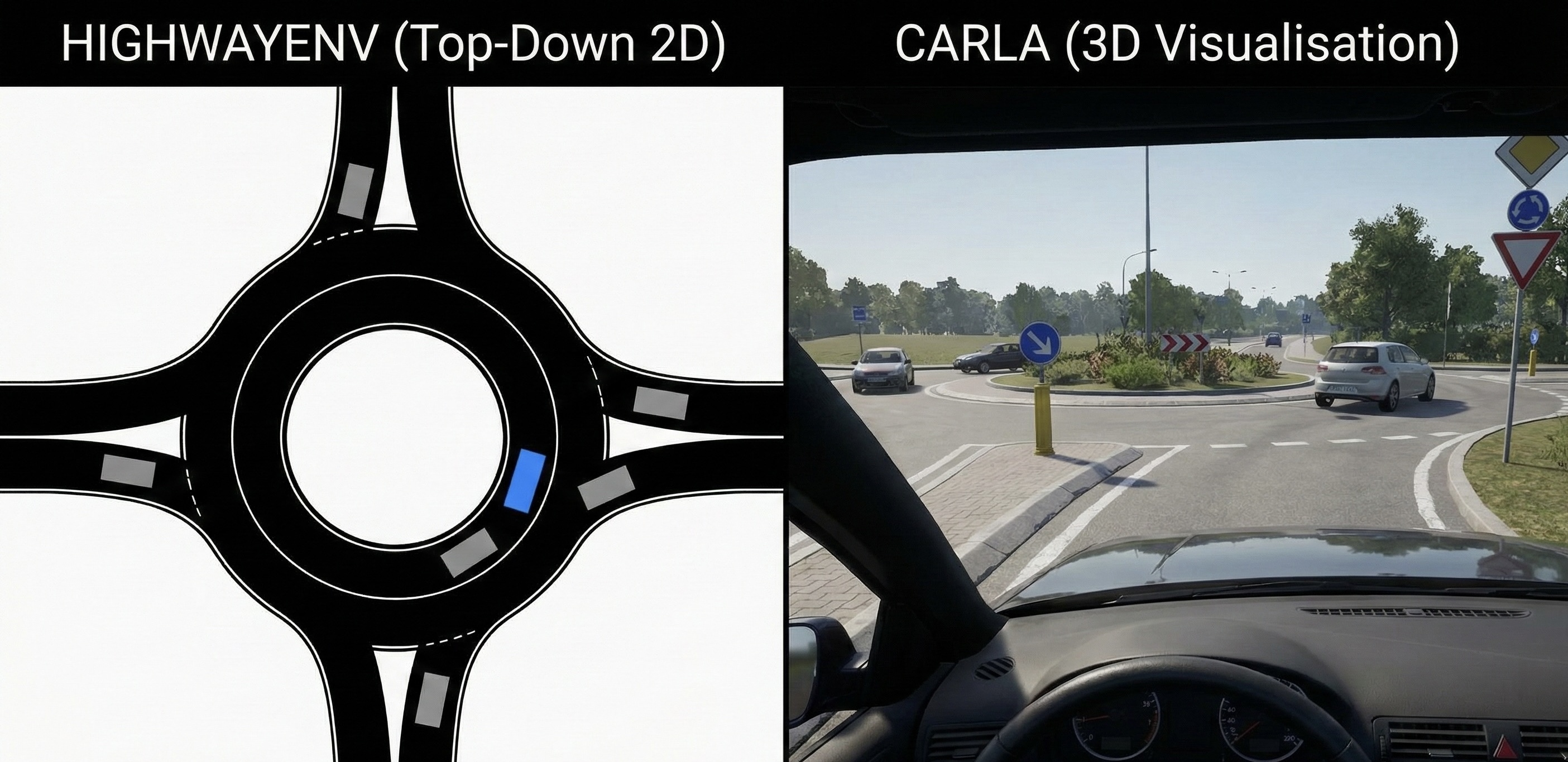}
    \caption{Roundabout.
    \textbf{Prompt:} \textit{"ego vehicle navigates
a roundabout"}}
    \label{fig:scenario_roundabout}
\end{subfigure}
\begin{subfigure}[b]{0.49\columnwidth}
    \includegraphics[width=\textwidth]{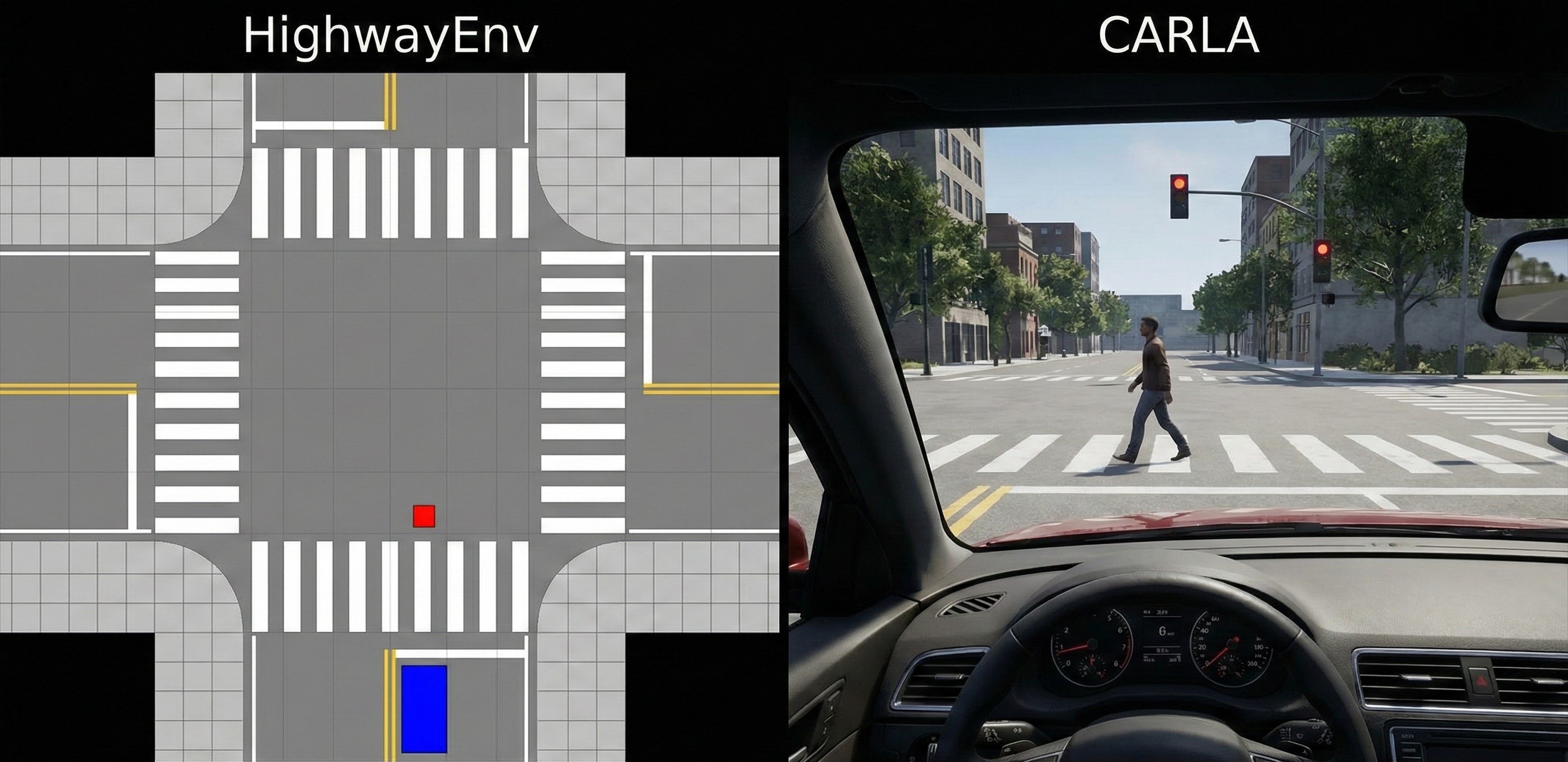}
    \caption{Pedestrian yielding. \textbf{Prompt:} \textit{"ego vehicle approaches
an urban intersection with a pedestrian crossing and correctly yields to a pedestrian at the
crosswalk"}}
    \label{fig:scenario_pedestrian}
\end{subfigure}
\hfill
\begin{subfigure}[b]{0.49\columnwidth}
    \includegraphics[width=\textwidth]{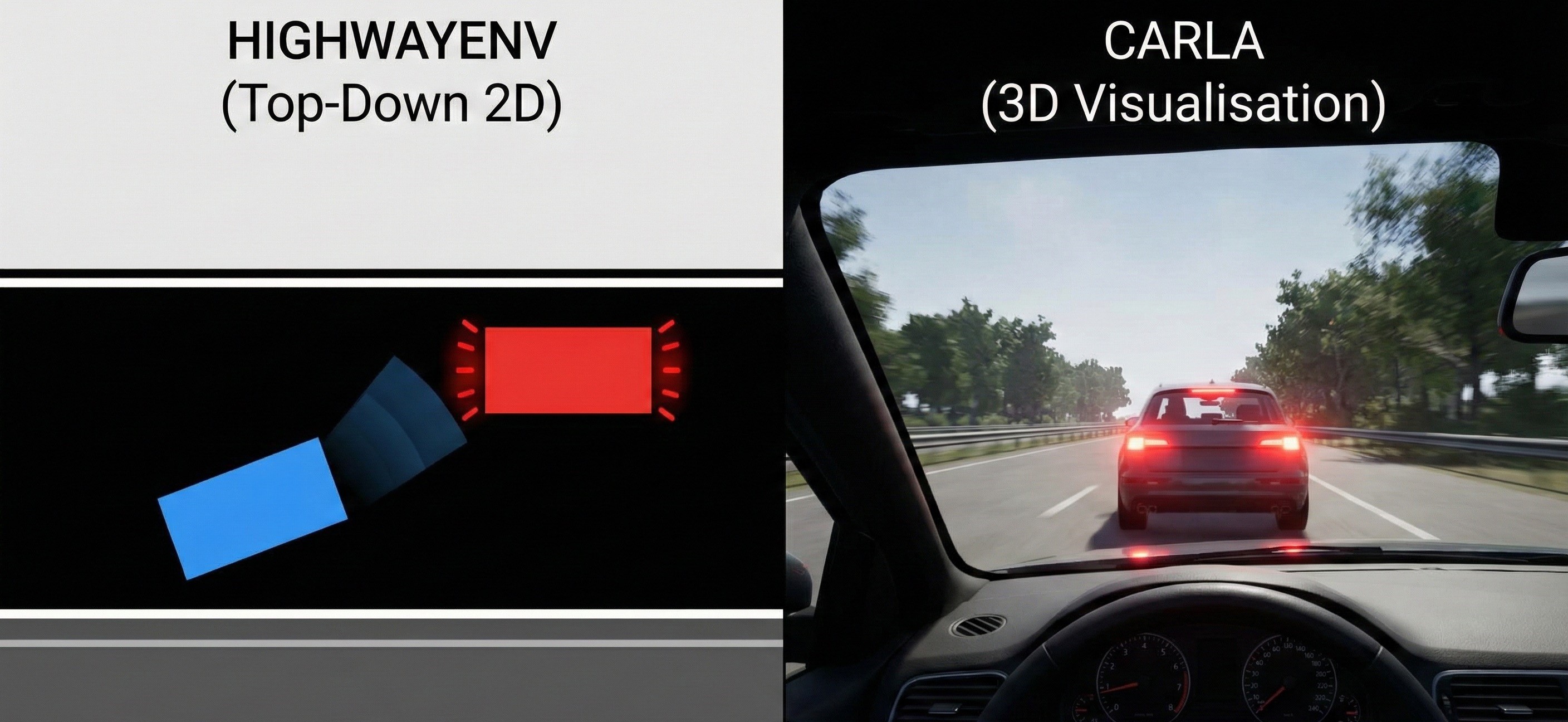}
    \caption{Failure case. \textbf{Prompt:} \textit{"ego vehicle crashes into another vehicle" }}
    \label{fig:failure}
\end{subfigure}

\caption{\textbf{State-interface transfer to CARLA.} ERQ-Net operates on highway, on-ramp, roundabout, and pedestrian scenarios through the shared graph-state representation. The final example shows a failure under abrupt adversarial braking.}
\label{fig:transfer_scenarios}
\vspace{-1em}
\end{figure}

\begin{figure}[!t]
\centering
\begin{minipage}[t]{0.48\textwidth}
    \vspace{0pt}
    \centering
    \vspace{-0.2cm}
    \begin{tabular}{lcc}
    \toprule
    \textbf{Condition} & \textbf{Succ.} & \textbf{Consistency} \\
    \midrule
    Clear & 73.1\% & Baseline \\
    Heavy rain & 72.8\% & 99.6\% \\
    Dense fog & 71.9\% & 98.4\% \\
    \midrule
    \textbf{Average} & \textbf{72.6\%} & \textbf{99.0\%} \\
    \bottomrule
    \end{tabular}

    \caption{\textbf{CARLA transfer across rendering conditions.} Success and trajectory consistency remain stable across clear, rain, fog, and night settings. Since ERQ-Net receives kinematics, the result measures graph-interface invariance rather than visual robustness.}
    \label{fig:weather_robustness}
\end{minipage}
\hfill
\begin{minipage}[t]{0.48\textwidth}
    \vspace{0pt}
    \centering
    \includegraphics[width=0.48\textwidth]{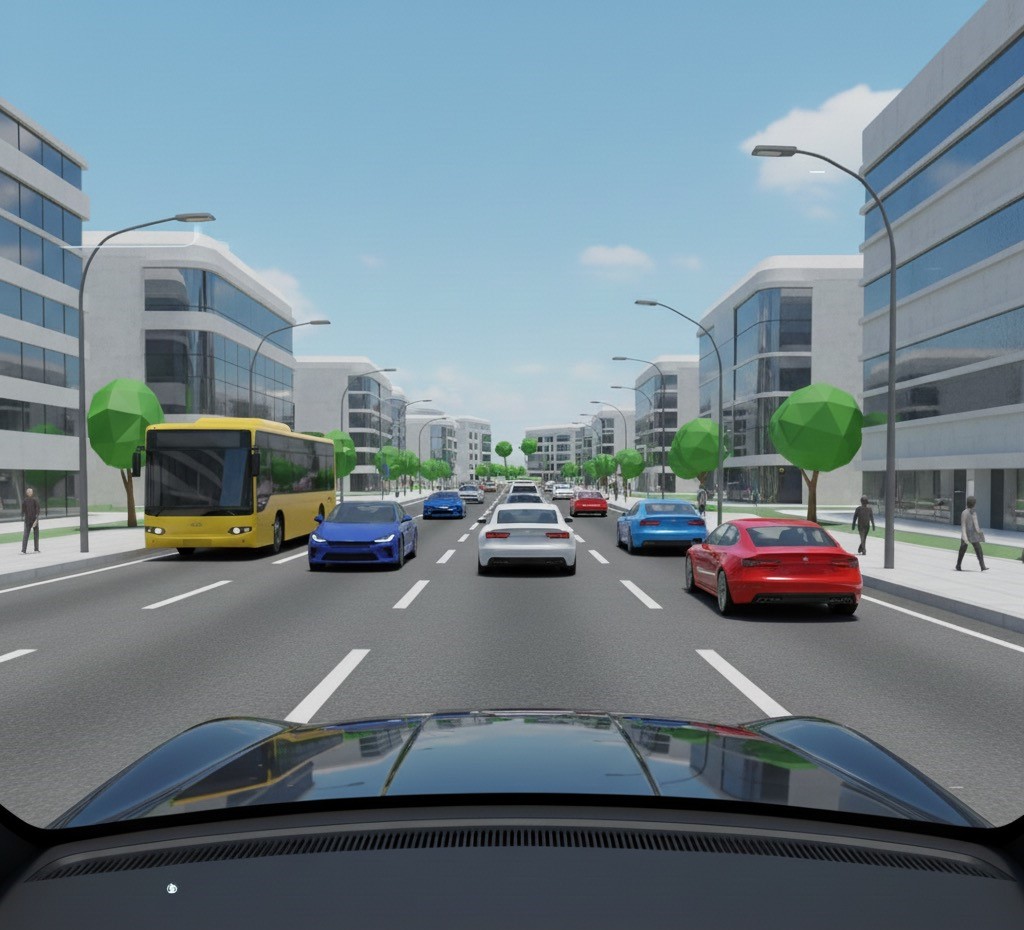}
    \hfill
    \includegraphics[width=0.48\textwidth]{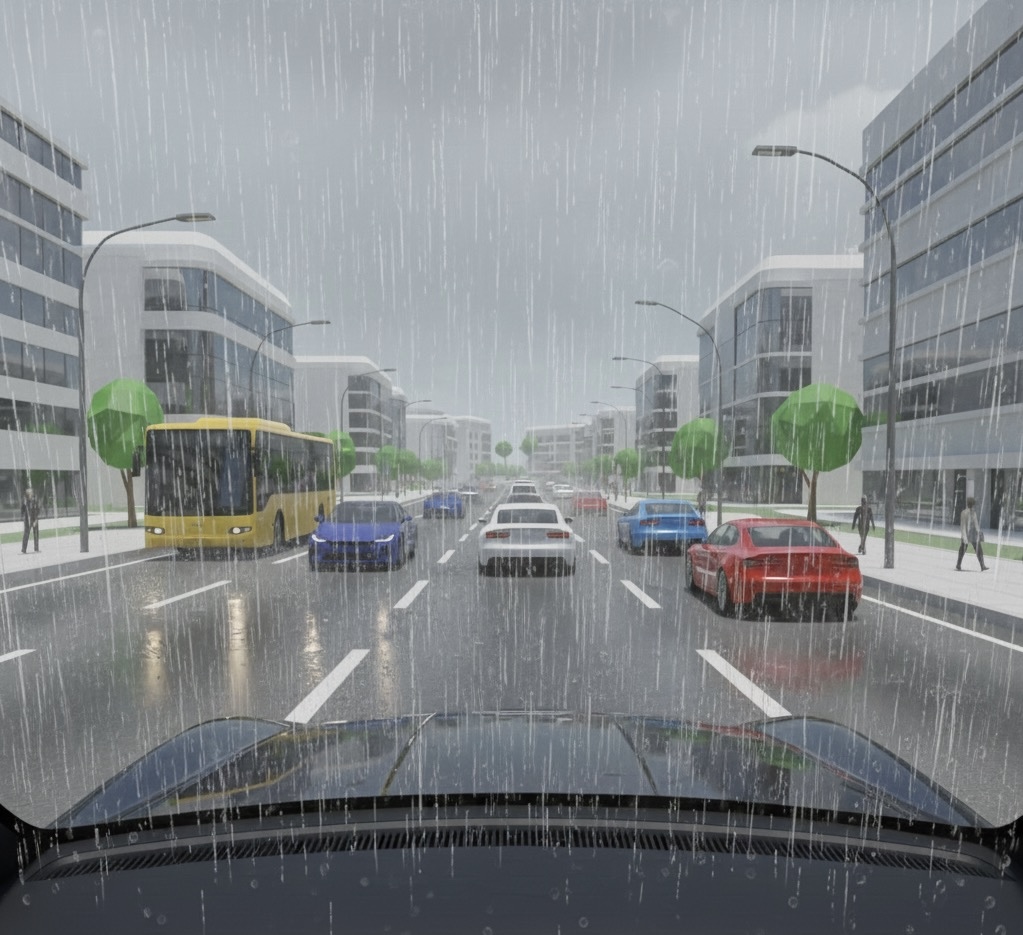}

    \vspace{0.1cm}

    \includegraphics[width=0.48\textwidth]{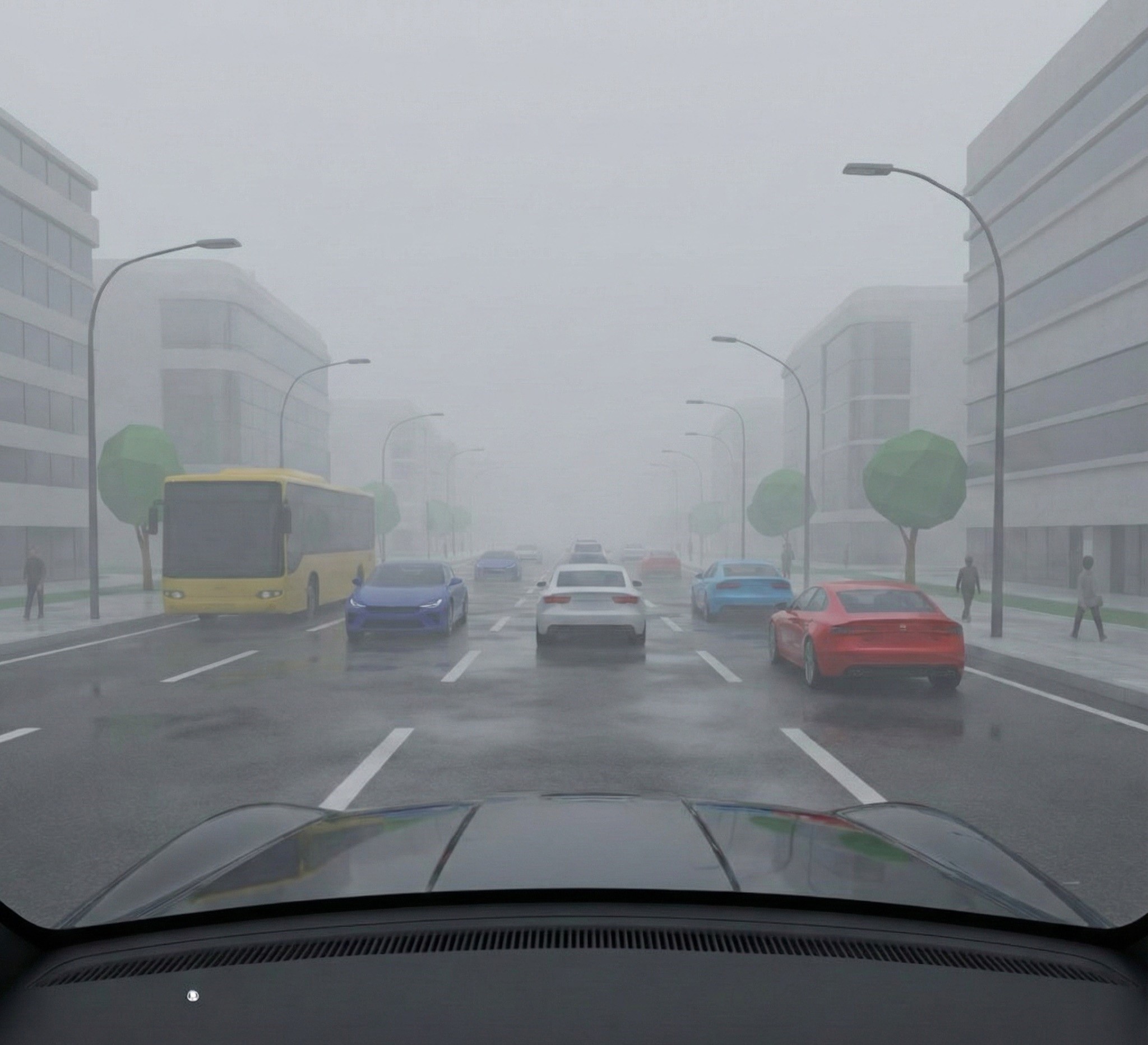}
    \hfill
    \includegraphics[width=0.48\textwidth]{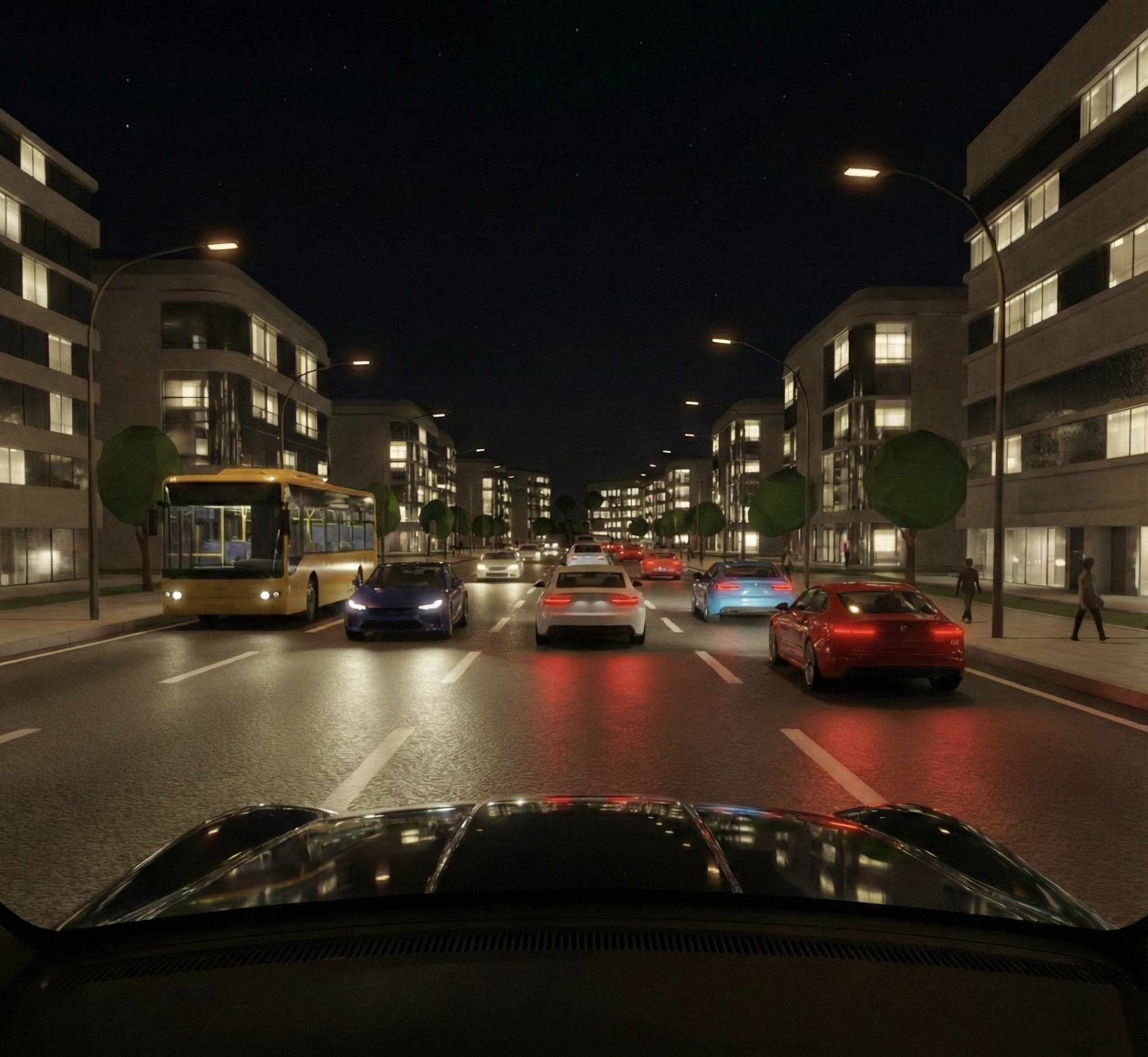} 
\end{minipage}
\vspace{-1em}
\end{figure}
Table~\ref{tab:cat} positions the proposed pipeline relative to CAT without treating the two systems as directly comparable. CAT perturbs real-log trajectories, whereas our framework provides log-free, language-level scenario specification and instantiates interactions through executable behaviour programs. CAT has lower per-scenario generation cost; our reported 1.2\,s is dominated by the one-time language-model parse, after which the configuration can be instantiated repeatedly. We thus treat the comparison as a difference in scenario-generation capabilities, not as evidence of statistical superiority.
\vspace{-.8em}
\subsubsection{State-Interface Transfer to CARLA.} 
\label{sec:transfer} 
Figure~\ref{fig:weather_robustness} evaluates ERQ-Net without fine-tuning through the common graph-state interface. On the dedicated transfer evaluation, CARLA achieves an average success rate of 72.6\%, 2.7 percentage points below the corresponding HighwayEnv result. Success remains stable under clear, heavy-rain, and dense-fog rendering, with 99.0\% mean trajectory consistency. Because ERQ-Net consumes simulator-derived kinematics rather than images, this consistency demonstrates invariance of the graph-state interface to rendering changes, not visual-perception robustness. The qualitative examples in Fig.~\ref{fig:transfer_scenarios} additionally show operation on ramps, roundabouts, and pedestrian interactions. These cases provide evidence of geometric portability, but do not constitute benchmark-scale out-of-distribution or real-world validation.
\vspace{-.8em}
\subsection{Scope and Future Work} 
\label{sec:limitations} 
The present study focuses on four types of controlled highway interaction and uses simulator-derived kinematic observations with a discrete 1\, Hz action space. This setting enables a reproducible analysis of language-structured relational learning but does not yet cover perception uncertainty, complex urban interactions, or continuous vehicle control. Future work will extend ERQ-Net to multimodal perception, higher-frequency and continuous control, and more compositional multi-actor scenarios. The observed recognition--control gap also motivates objectives that better preserve scenario-dependent action diversity.
\vspace{-.8em}
\section{Conclusion}
\label{sec:conclusion}
We introduced Language-Structured Relational Q-Learning, instantiated through ERQ-Net, for threat-aware control in safety-critical driving. ERQ-Net couples dynamic graph-based interaction reasoning with value-based action selection, allowing actor relevance, ego-centric representation, and action values to be learned jointly. Language structures the surrounding actors' behaviour during training, while prompts and semantic actor labels remain hidden from the policy. 

Across 2,500 scenarios, language-structured training improves success by up to six percentage points over the matched random control and increases adversary-focused attention from $1.2\times$ to $2.1\times$. Relational-encoder ablations further confirm the benefit of graph structure and learned interaction relevance. However, trained policies remain comparable to the best constant action, despite a portfolio of simple policies that solve 76\% of the test scenarios. This 18-point recognition--control gap shows that identifying the relevant threat does not necessarily produce the appropriate scenario-dependent response. The CARLA experiments additionally demonstrate zero-shot state-interface transfer without fine-tuning. Overall, the results establish both the value of language-structured relational learning and the need to evaluate whether emergent threat awareness is genuinely converted into adaptive control.


\FloatBarrier 

\bibliographystyle{splncs04}
\bibliography{references}
\end{document}